\documentclass{article}
\usepackage{amsmath,graphicx}
\usepackage{spconf}
\usepackage[symbol]{footmisc}

\usepackage{xcolor}
\usepackage{soul}

\begin{document}
\title{Improving Feature Generalizability with Multitask Learning in Class Incremental Learning}
%
\name{Dong Ma$^{\dag,1,2}$, Chi Ian Tang$^{\dag, 1}$\thanks{$\dag$ Authors with equal contribution in alphabetical order.}, Cecilia Mascolo$^1$}
%
\address{$^1$University of Cambridge, $^2$Singapore Management University \\ 
\{dm878, cit27, cm542\}@cam.ac.uk}
%
%

\ninept
\maketitle
\begin{abstract}

Many deep learning applications, like keyword spotting~\cite{chen2014small,szoke2005comparison},
require the incorporation of new concepts (classes) over time, referred to as Class Incremental Learning (CIL). The major challenge in CIL is 
catastrophic forgetting, i.e., preserving as much of the old knowledge as possible while learning new tasks. Various techniques, such as regularization, knowledge distillation, and the use of exemplars, have been proposed to resolve this issue. However, prior works primarily focus on the incremental learning step, while ignoring the optimization during the base model training. We hypothesise that a more transferable and generalizable feature representation from the base model would be beneficial to incremental learning. 

In this work, we adopt multitask learning during base model training to improve the feature generalizability. Specifically, instead of training a single model with all the base classes, we decompose the base classes into multiple subsets and regard each of them as a task. These tasks are trained concurrently and a shared feature extractor is obtained for incremental learning. We evaluate our approach on two datasets under various configurations. The results show that our approach enhances the average incremental learning accuracy by up to 5.5\%, which enables more reliable and accurate keyword spotting over time.
Moreover, the proposed approach can be combined with many existing techniques and provides additional performance gain.

\end{abstract}
\begin{keywords}
Class Incremental Learning, Continual Learning, Multitask Learning, Keyword Spotting
\end{keywords}
\section{Introduction}
\label{sec:intro}
Recently, deep learning has enabled the boom of a variety of applications such as face recognition~\cite{wang2014face} and keyword spotting~\cite{chen2014small,szoke2005comparison}. Albeit remarkable performance, deep learning models are usually built upon a fixed dataset, lacking the ability and flexibility of adapting to sequentially incoming data (of the same class or new classes)~\cite{diethe2019continual}. For example, in keyword spotting, the initial model is built based on a pre-defined keyword set. When the user intends to add new keywords over time, the data-hungry nature of deep learning incurs two challenges for model update. First, storing the previous data might require substantial memory. Second, retrieving old data and training a new model from scratch requires considerable time and computational resources. A research area, class incremental learning (CIL), which tries to retain the acquired knowledge while learning new concepts, has been initiated to address these challenges~\cite{van2019three}. 

\begin{figure}
    \begin{center}
        \includegraphics[width=0.46\textwidth]{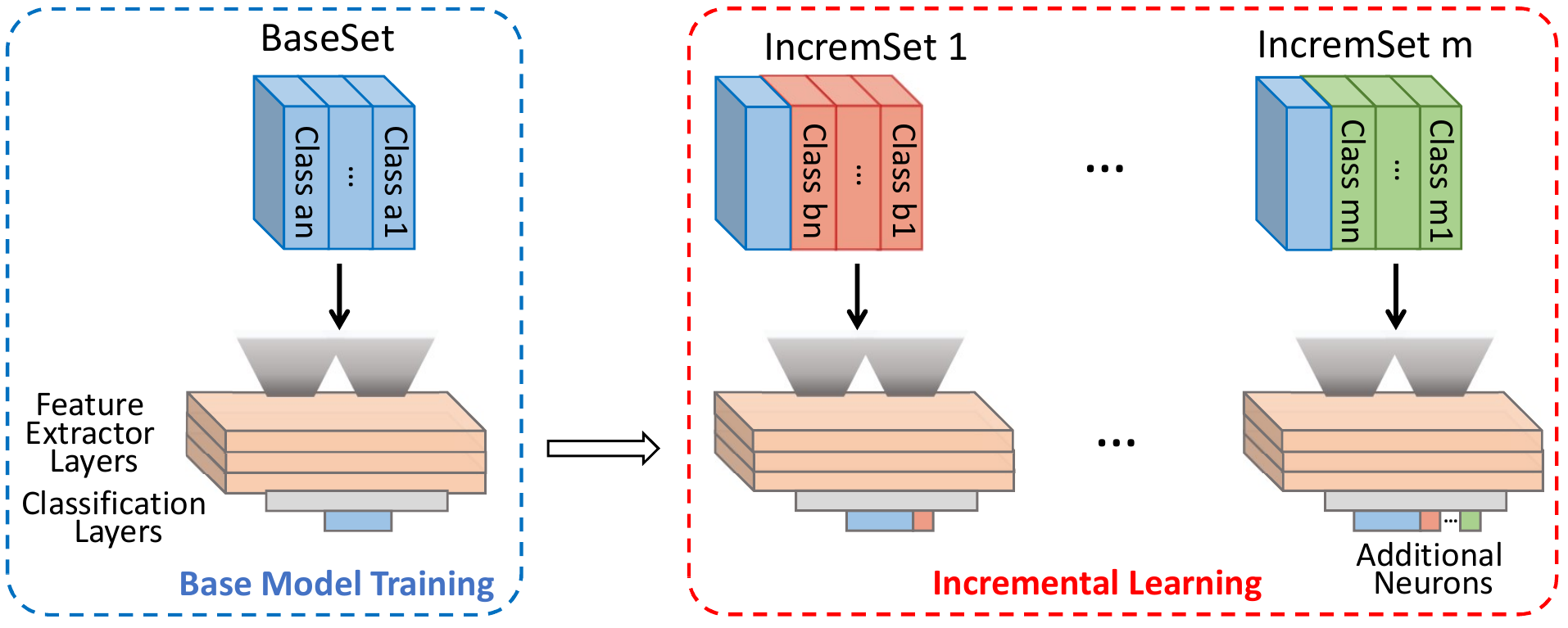}
    \vspace{-0.15in}
    \caption{Illustration of typical class incremental learning.}
    \vspace{-0.3in}
    \label{figure:cil_overview}
    \end{center}
\end{figure}

Typically, CIL consists of two stages, as shown in Figure~\ref{figure:cil_overview}. The base model training stage utilizes the full dataset at hand to train a base model. During the incremental learning stage, the base model will be fine-tuned with new data received over time (and a part of the old data). A critical problem in CIL is {\it catastrophic forgetting}~\cite{diethe2019continual,van2019three,aljundi2018memory,kirkpatrick2017overcoming,zenke2017continual,li2017learning}, that is, the model tends to be overfitted on the new class data, while forgetting previously acquired knowledge, thereby degrading the inference performance on old classes. This originates from the imbalance between the old and new class data as none or only part of the old data is retained to save storage during fine-tuning~\cite{wu2019large,hou2019learning}. Moreover, the data imbalance issue also implies  bias at the classification layer: old samples are more likely to be classified as new classes~\cite{zhao2020maintaining}. 

To deal with catastrophic forgetting, researchers have proposed different techniques, which can be divided into three categories. The initial attempt to resolve catastrophic forgetting is regularization. Specifically, this method first locates the important model parameters that contribute more to the final classification by using certain metrics such as the fisher information~\cite{kirkpatrick2017overcoming} and output gradients~\cite{aljundi2018memory}. Then, an additional term is applied to the loss function to restrict the changes of these important weights. Later, knowledge distillation~\cite{hinton2015distilling}, which transfers knowledge between different domains or networks, has been adopted to retain the old knowledge when learning a new task~\cite{li2017learning,zhou2019m2kd}. The output logits of new samples on the old model will be recorded first, and an extra loss term is added to minimize its difference with the output logits from the new model during incremental training. These two approaches have been proven to have poor performance in CIL~\cite{jha2021continual}. Most recently, exemplar-based approaches were proposed, where a small portion of old samples are selected (randomly or with herding technique~\cite{lopez2017gradient,hayes2020remind,iscen2020memory}) and stored. In the incremental learning step, the exemplar set is combined with new data for model updates, which shows great performance in retaining old knowledge and learning new concepts. 
The state-of-the-art incremental learning performance is achieved with a combination of different techniques~\cite{mittal2021essentials}.

However, we identified that {\em all the existing approaches focus on the incremental learning step, while ignoring the optimizations in the base model training stage}. Intuitively, to retain the old knowledge while learning new concepts, we would that expect the feature representation (or embedding) produced by the base model to be general to the new classes as well~\cite{mittal2021essentials}. In other words, if we can train a more representative and transferable base model, we can alleviate the catastrophic forgetting issue during incremental learning. 
This assumption is tenable due to the fact that for a given neural network and accuracy, there exist multiple sets of weights, i.e., training a neural network multiple times with the same setting results in different sets of model weights. So, the question is \textit{how to obtain a set of weights that is more transferable to new classes}.

In this work, {\em we propose the use of multitask learning during the base model training to improve the generalizability of embeddings}. Multitask learning involves learning multiple tasks in parallel while using a shared representation~\cite{caruana1997multitask}. If we define different subsets of base classes as different classification tasks, the model would see different combinations of the base classes. As a result, the model is forced to learn a set of weights based on different views of the dataset, which would be more transferable to new classes. Moreover, the number of training epochs needed for incremental learning is decided empirically in current approaches. We introduce a validation set and the early stopping technique to avoid underfitting or overfitting in practical scenarios. Experimenting with the Google Speech Commands (GSC) and UrbanSound8K datasets, we demonstrated that our approach can further enhance the average classification accuracy by up to 5.5\%, which enables more reliable and accurate keyword spotting over time. 
Furthermore, our approach is compatible with many other existing approaches and adds extra performance gains over them as it works on the base model training stage.


\section{Primer}
\label{sec:format}
\subsection{Related Work}
Existing literature on CIL mainly addresses two research problems, i.e., catastrophic forgetting of old knowledge and data imbalance between old and new classes.  

\textit{Catastrophic forgetting}: refers to the fact that the model forgets previously attained knowledge on old classes when learning new concepts with new class data. Previous works to combat catastrophic forgetting can be categorized into two groups - without and with samples of old classes. Techniques such as parameter control~\cite{aljundi2018memory,kirkpatrick2017overcoming,zenke2017continual}, knowledge distillation~\cite{hinton2015distilling}, exemplar replay~\cite{lopez2017gradient,hayes2020remind,iscen2020memory,liu2020mnemonics,castro2018end,rebuffi2017icarl}, and generative adversarial network~\cite{goodfellow2014generative} are proposed in the literature, as discussed in the Introduction.

\textit{Data imbalance}: no matter with or without old samples, there exists a huge imbalance between old and new samples during incremental learning stage. As a result, the model will be overfitted to the new classes. To address it, BiC~\cite{wu2019large} adds an extra bias correction layer to correct
the model’s outputs. WA-MDF~\cite{zhao2020maintaining} aligns the norm of new class weight vectors to
that of the old class weight vectors and~\cite{hou2019learning} applied cosine normalization in the classification layer.

Overall, the state-of-the-art approaches combine multiple techniques such as knowledge distillation and data balancing to achieve optimal performance~\cite{mittal2021essentials}.  

\subsection{Motivation}
For each line of work, we identified certain limitations that motivate this work. 

First, existing approaches that deal with catastrophic forgetting all focus on the incremental learning stage, while ignoring the importance of the base model training. In particular, we observed that training a neural network for multiple times under the same setting\footnote{Without changing any hyper-parameter and simply run the same code again to achieve a similar accuracy.} will result in different sets of weights (or embeddings), i.e., the model weights are not unique. This observation raises a question, \textit{among different sets of model weights, which one is more beneficial for incremental learning?} Since less transferable embeddings will force the model to alter its weights significantly to learn new concepts, the previously attained knowledge will be destroyed, leading to more severe catastrophic forgetting. Thus, we hypothesise that \textit{it is possible to alleviate catastrophic forgetting by training a more transferable base model.}

\begin{figure}[t]
\begin{minipage}[b]{.32\linewidth}
  \centering
  \centerline{\includegraphics[height=2.5cm]{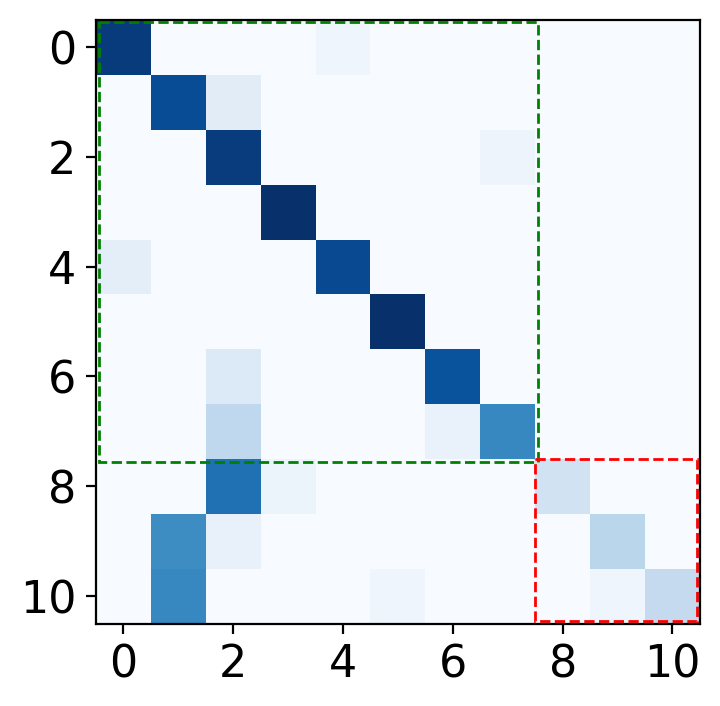}}
  \centerline{(a) Epoch 15}
\end{minipage}
\hspace{-0.1in}
\begin{minipage}[b]{.32\linewidth}
  \centering
  \centerline{\includegraphics[height=2.5cm]{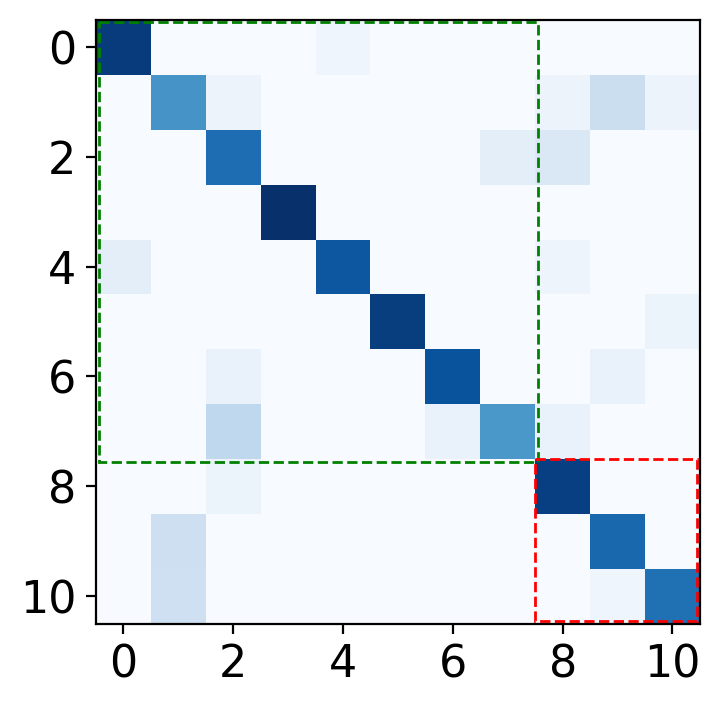}}
  \centerline{(b) Epoch 45}
\end{minipage}
\hspace{0.03in}
\begin{minipage}[b]{.32\linewidth}
  \centering
  \centerline{\includegraphics[height=2.5cm]{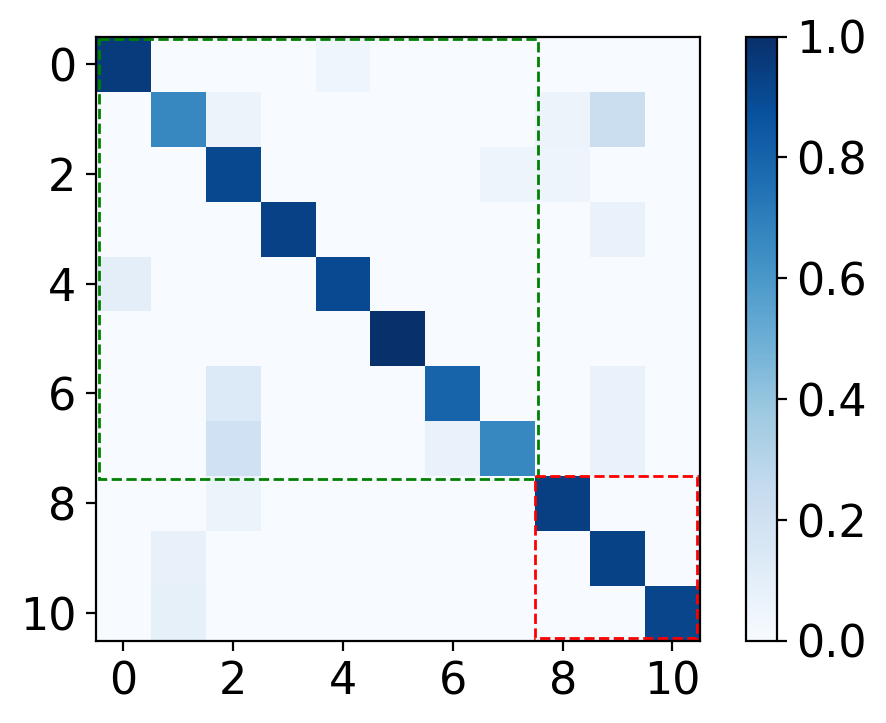}}
  \centerline{(c) Epoch 100}

\end{minipage}
\vspace{-0.1in}
\caption{Comparison of confusion matrix after training for (a) 15 (acc old = 0.91, acc new = 0.24), (b) 45 (acc old = 0.83, acc new = 0.82), and (c) 100 (acc old = 0.86, acc new = 0.92) epochs. The green and red box represent old and new classes, respectively. }
\vspace{-0.25in}
\label{fig:confusion_matrix}
\end{figure}

Second, during incremental learning, existing works usually report an empirical value for training epochs\footnote{Such empirical value is obtained with the full dataset of the new classes, which is unavailable in practical CIL setting.} and this number varies a lot for the same dataset. For instance, on the CIFAR-100 datset, the number of training epoch for BiC~\cite{wu2019large}, WA-ADB~\cite{zhao2020maintaining} and iCarl~\cite{rebuffi2017icarl} is set to 250, 180, and 70 respectively. With the increase of training epochs, the model tends to shift to new classes due to data imbalance between old classes and new classes. Figure~\ref{fig:confusion_matrix} compares the confusion matrix of a CIL task at different training epochs, where class 0-7 are old classes and class 8-10 are new classes. We can observe that when the number of training epoch is small (i.e., 15), the model has not acquired enough knowledge about the new classes and the accuracy on new classes is poor. With more training epochs, the performance on new classes increases gradually, and the model becomes overfitted with excessive epochs (i.e., 100). Consequently, setting a proper value for training epochs is critical to deal with data imbalance in CIL. 


\section{Methodology}
\label{sec:pagestyle}
\begin{figure}
    \begin{center}
        \includegraphics[width=0.46\textwidth]{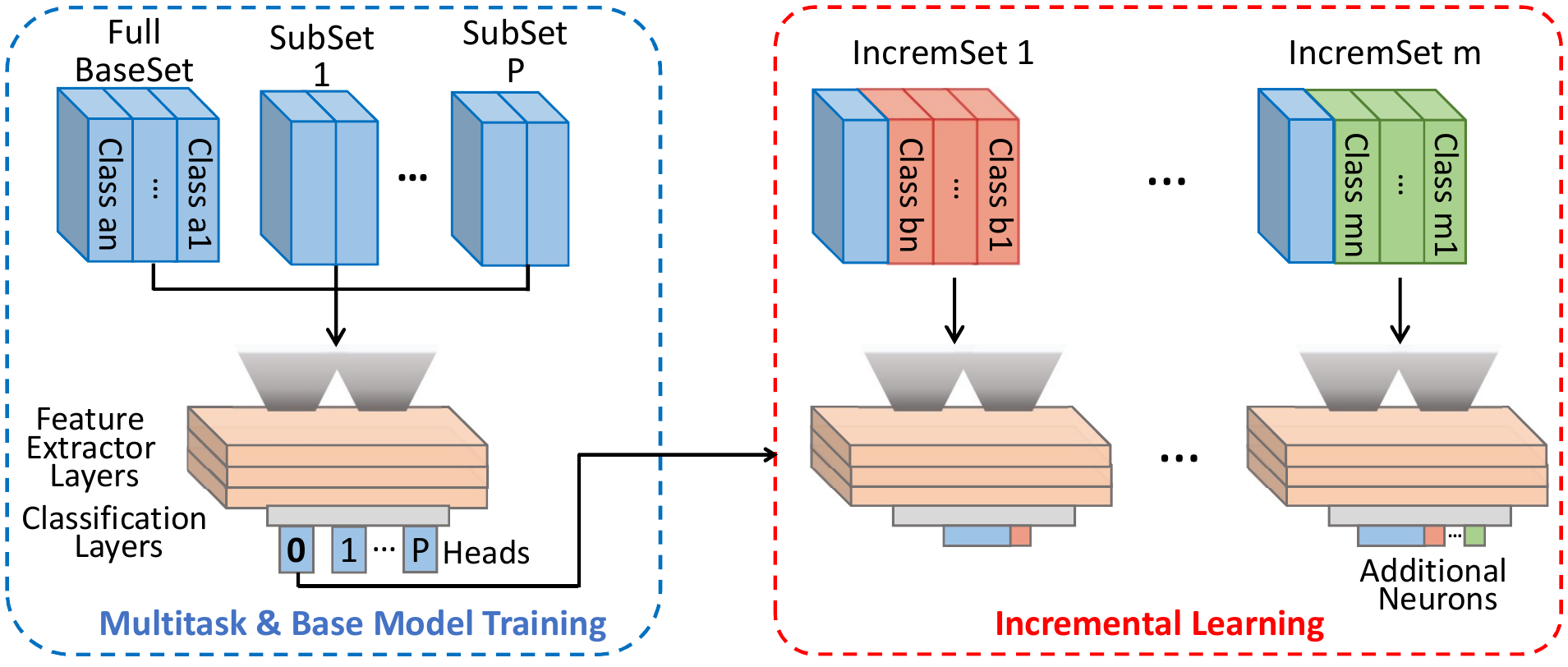}
    \vspace{-0.1in}
    \caption{Illustration of multitask learning based CIL.}
    \vspace{-0.3in}
    \label{figure:multitask_cil}
    \end{center}
\end{figure}

Motivated by the discussion above, we introduce \textit{multitask learning} to CIL to improve the generalizability of feature representations. As shown in Figure \ref{figure:multitask_cil}, the full base dataset is decomposed into multiple subsets and they form a multitask setting, where the classification of each set is regarded as a task. These tasks are trained concurrently with a shared representation. After multitask training, the model backbone and the largest head (corresponding to full base dataset) is forwarded to the incremental learning step, where the state-of-the-art techniques can be directly applied.

\subsection{Intuition of Multitask Training}
Multitask learning forces the model to simultaneously solve multiple tasks at once~\cite{caruana1997multitask}, and has been used to help improve the generalization of the model in solving all tasks, by preventing the model from overfitting to any particular task \cite{multi_task_survey_9392366}. In the context of CIL, generalizability of the model is crucial, since it is necessary for the model to retain previous knowledge while learning about new classes. We designed a multitask learning scheme that aims to simulate the incremental learning steps at the base model training stage, by dividing the dataset into different subsets.
For example, with base classes such as `air conditioner', `car horn', `children playing' and `drilling', we can create different classification tasks by taking clusters
of them, including \{`air conditioner', `car horn'\}, \{`air conditioner', `drilling'\} and \{`air conditioner', `car horn', `children playing', `drilling'\}. This grouping of the base learning classes aims to make the model find a solution which can solve all these classification tasks at once, instead of overfitting to the whole base dataset. As a result, the generalizability of the base model is improved, leading to better performance in the incremental learning stage. 
This particular setup is analogous to the incremental learning stage, where the model is required to perform well on different sets of classes.

\subsection{Multitask Creation}\label{sec:multitask}
Important design choices that arise from adopting multitask training are: (1) the number of tasks, and (2) the cluster of classes that each task corresponds to. If we have $N$ target classes in the dataset used in base model training, there are $2^N-1$ distinct valid tasks (which equals to the number of subsets excluding the empty set). With the number of task being 1 and the cluster being the entire set of classes, 
we have the conventional setup of a fully-supervised training on the base dataset. The number of distinct tasks grows very rapidly as $N$ increases, so it is infeasible to train on all of them, and a design choice should be made. With the number of classes being closer to $N$, the task itself is harder. However, restricting all the tasks to have high number of classes reduces the diversity of tasks presented to the model, and it may not help in improving the generalizability of the solution that the model converges to. In this work, we explore different choices of tasks, along two directions: (1) tasks with different number of classes (such as 5, 4, and 3 classes for each task), and (2) those with different subsets of classes but the same number of classes (such as 5, 4, 4 and 4 classes, but each task is a distinct subset). 

\subsection{Learning Rate Scheduler}
One of the most important hyperparameters affecting the amount of knowledge retention in CIL is the learning rate during incremental steps, in which the models are fine-tuned. Some existing works adopts a cosine annealing for training with a large starting learning rate ($0.01$)~\cite{mittal2021essentials}. We observed that this large starting learning rate significantly changes the weights of the neural network, and this could make knowledge retention difficult. On the other hand, adopting a fixed small learning rate causes training to be slow. 

As a result, we propose a two-step fine-tuning strategy to mitigate this. First, the majority of the up-stream layers of the neural network are frozen, while the down-stream layers which include the classification layer with the newly added neurons are trained with a relatively large learning rate. After the down-stream layers, and the new neurons in particular, converge to a reasonable solution to both new and old classes, the up-stream layers are unfrozen and the entire network is fine-tuned with a small learning rate. An early stopping mechanism is also adopted in our scheme because it is difficult to find the balance between overfitting and underfitting when using a fixed number of training epochs (which is commonly adopted in previous works). 


\section{Evaluation}
\label{sec:typestyle}
\subsection{Experiment Setting}
\textbf{Dataset}: We evaluated the proposed approach on two audio datasets.  
UrbanSound8K dataset~\cite{salamon2014dataset} consists of 10 different environment sound events such as drilling, car horn, street music, etc. There are 8,732 audio clips sampled at 22~kHz, where each clip lasts for 3-4 seconds. Following~\cite{kwon2021fasticarl}, we extracted four audio features (Log-mel spectrogram, chroma, tonnets, and spectral contrast) using the first 3-seconds of a clip. As a result, the created input has a size of 128$\times$85, where 128 represents the number of frames and 85 represents aggregated feature size of the four audio features. Google Speech Commands (GSC)~\cite{salamon2014dataset} is a widely used dataset in keyword spotting. We pick the 20 core keywords as the classes and each class contains 3,200 1-second clip. Log-mel spectrogram features have been extracted and a 32$\times$32 input is created. 

For UrbanSound8K dataset, we set the number of base class and incremental class as 4 and 2 respectively, i.e., 4-2-2-2. Similarly, GSC dataset is split to 5-3-3-3-3-3. For both datasets, the (train, test, validation) splitting ratio is set to (0.7, 0.2, 0.1). We allocate a fixed memory to store $K$ (default 100 for UrbanSound8K and 200 for GSC) exemplars for all previously seen classes. 

\noindent\textbf{Model Architecture: }We designed a convolutional neural network to classifier the audio events. The network consists of four Conv2D layers with ReLU activation, each followed by a BatchNormalization layer. Then, a Dropout (0.5) layer and an AveragePooling layer is connected before the final classification layer. Stochastic gradient descent is used as the optimizer. All the code is implemented in PyTorch and run on a NVIDIA GeForce RTX 2080 GPU. We use the hard parameter sharing approach for our multitask learning, where the hidden layers for different tasks are shared and remain exactly the same and only the classification layer is trained separately.

\noindent\textbf{Baseline: }We compare our work to the method proposed by Mittal et al. \cite{mittal2021essentials} which reported state-of-the-art results for CIL. Specifically, \cite{mittal2021essentials} first utilizes the cross entropy (CE) loss and knowledge distillation (KD) loss on new classes to learn new knowledge. Then, it constructs a small but balanced exemplar set (including current incremental classes) to correct the bias and preserve old knowledge (with CE loss and KD loss). For more details, please refer to the original paper~\cite{mittal2021essentials}. We adopted the same incremental learning strategy as in~\cite{mittal2021essentials} and the difference only comes from the base model training, i.e., single task (\cite{mittal2021essentials}) vs. multitask (ours).

\subsection{Results}

\begin{figure}[t]
{\centering
\includegraphics[width=0.95\linewidth]{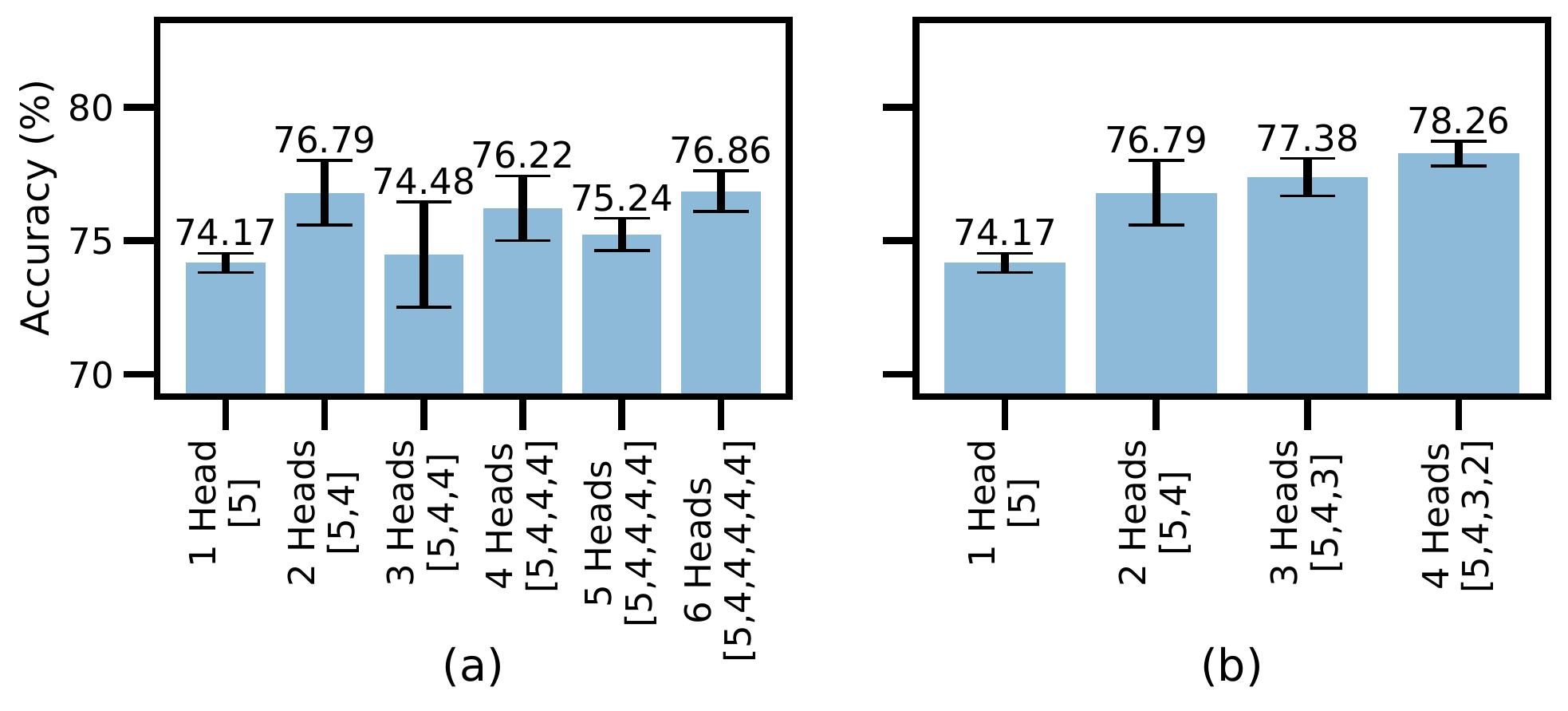}
}

  
%
\vspace{-0.2in}
\caption{Impact of task creation. }
\vspace{-0.2in}
\label{fig:task_creation_performance_decreasing}
\end{figure}

\begin{table}[]
\centering
\caption{Time overhead when training with different number and choice of tasks/heads.}
\vspace{0.05in}
\begin{tabular}{|c|c|c|c|}
\hline
Heads     & \begin{tabular}[c]{@{}c@{}}Training Time\\ per Epoch (s)\end{tabular} & Heads         & \begin{tabular}[c]{@{}c@{}}Training Time \\ per Epoch (s)\end{tabular} \\ \hline
[5]& 2.16  & [5,4,4]  & 4.79 \\ \hline
[5,4]& 3.41  & [5,4,4,4] & 5.35  \\ \hline
[5,4,3]  & 4.65  & [5,4,4,4,4]   & 6.44  \\ \hline
[5,4,3,2] & 5.44  & [5,4,4,4,4,4] & 7.42 \\ \hline
\end{tabular}
\vspace{-0.2in}
\label{tab:heads_time}
\end{table}

\noindent\textbf{Impact of Task Creation: }
First, we systematically explored the configurations of different tasks along two directions: (1) tasks with different number of classes, and (2) those with different subsets of classes. Figure \ref{fig:task_creation_performance_decreasing} demonstrates the change in model accuracy as we vary the configuration of tasks, on the GSC dataset. From Figure \ref{fig:task_creation_performance_decreasing}(a), we could infer that adding more tasks with the same number of classes but different subsets did not lead to consistent performance improvements. On the other hand, from Figure \ref{fig:task_creation_performance_decreasing}(b), we could see that increasing the number of tasks with different classes led to a consistent increase of performance, up to 5\%, although a diminishing effect could be seen when we add more tasks. 

We hypothesise that having many similar tasks may not offer much help in improving the model's generalizability, while having more distinct tasks is better at doing that. This is because our multitask training objective is to allow the model to see wider and more diverse combinations of the base classes, so that it can be effectively extended to new classes during incremental learning. Consequently, if the tasks are too similar (i.e., huge overlapping of classes), the model might not be able to learn more generalizability. For example, in the extreme case, if all the tasks contain the same classes, our approach degrades to single task learning. 

For a given set of base classes, a good construction of multiple tasks is critical for our approach. In this work, we focused on the number of classes and tasks in multitask learning, without utilizing the prior knowledge about the dataset. For example, grouping classes based on their semantic meaning might be helpful in creating high-quality tasks. We will explore it in the future.

In terms of time overhead, from the results shown in Table \ref{tab:heads_time}, we could observe consistent increase in training time per epoch when the number of tasks was increased, and when the number of classes in each task was increased.

Overall, the results demonstrated that having more than a single task helps in improving the model's performance, especially when the tasks are more distinct from each other. We will therefore select `4 heads [5, 4, 3, 2]' as the multitask setting for the rest of the evaluation.




\begin{figure}[t]
\begin{minipage}[b]{.49\linewidth}
  \centering
  \centerline{\includegraphics[width=4.0cm]{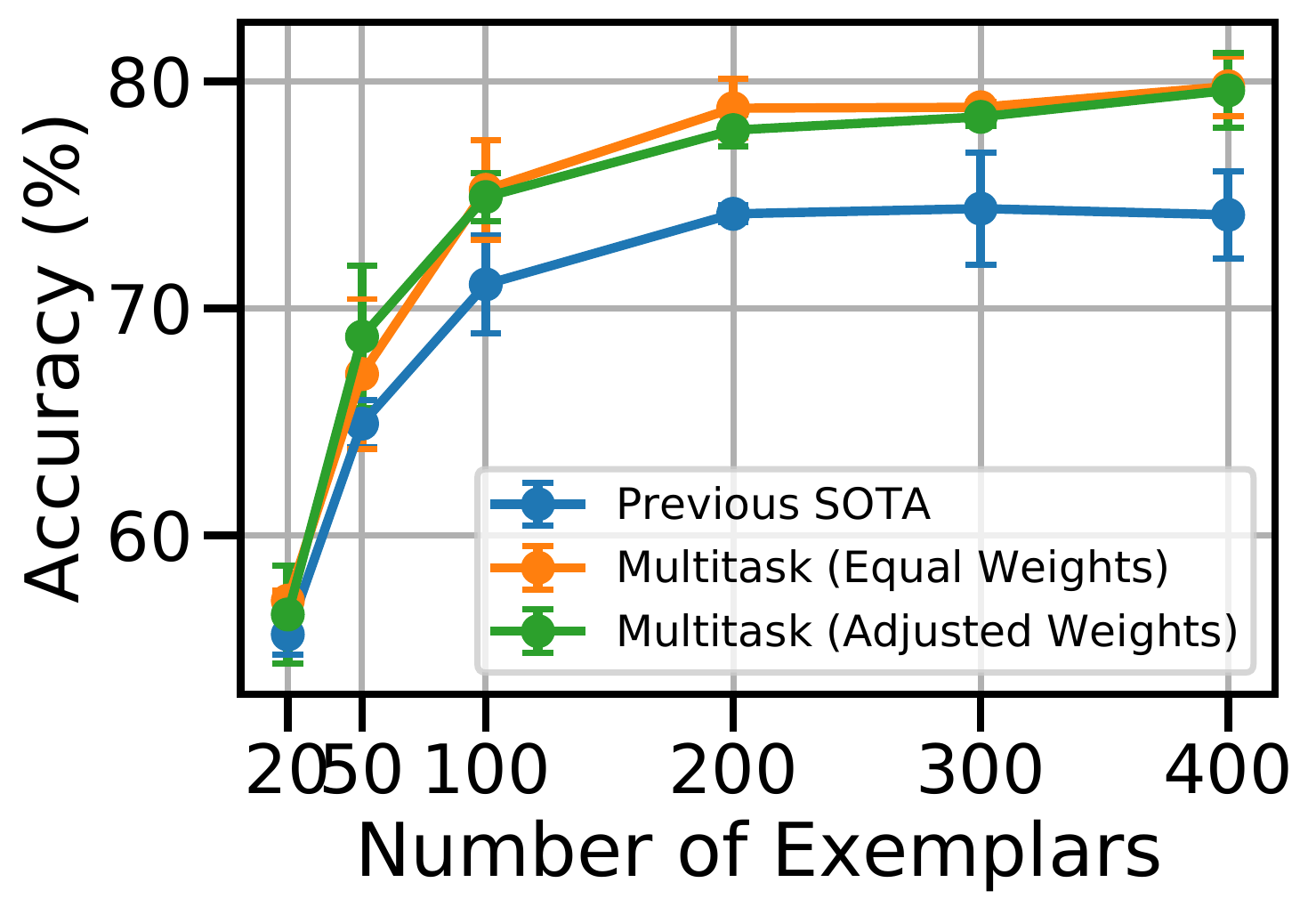}}
  \centerline{(a) GSC}
\end{minipage}
\hspace{-0.1in}
\begin{minipage}[b]{.49\linewidth}
  \centering
  \centerline{\includegraphics[width=4.0cm]{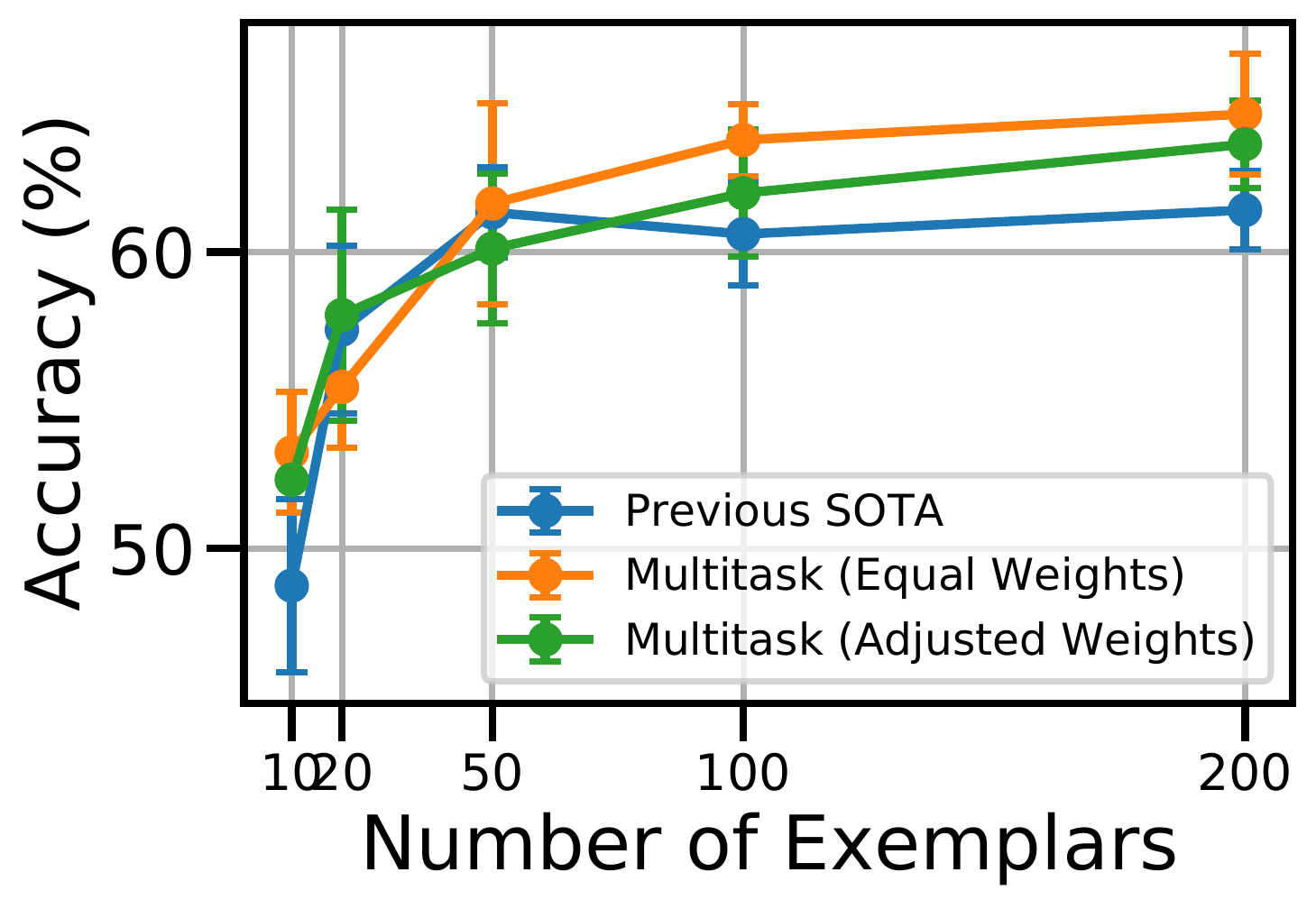}}
  \centerline{(b) UrbanSound8K}
\end{minipage}
\vspace{-0.1in}
\caption{Impact of exemplar quantity. }
\vspace{-0.25in}
\label{fig:num_exemplars}
\end{figure}



\noindent\textbf{Impact of Exemplar Quantity: }As the number of exemplars available at the incremental learning stage can have a significant impact on the performance, we further conducted a set of experiments with varying quantity of exemplars: 20, 50, 100, 200, 300 and 400 for Google Speech Commands (GSC) and 10, 20, 50, 100 and 200 for UrbanSound8K (US8K). Figure \ref{fig:num_exemplars} shows the accuracy of the model across different amounts of exemplars. We could see a general trend of improving performance as the number of exemplars was increased, with a diminishing effect. The performance of models starts to plateau with around 200 exemplars on the GSC dataset, and around 100 exemplars on US8K. 

\noindent\textbf{Comparison with Baseline: }To compare with the state-of-the-art method, we conducted the same set of experiments mentioned in the previous section, using only a single task as the baseline.
From Figure \ref{fig:num_exemplars}, we could see that our method started with similar performance as the previous state-of-the-art method when very few exemplars were allowed, and showed a consistent improvement of up to 5.5\% in accuracy compared to the baseline as the amount of exemplars increased on GSC, and up to 3.2\% on US8K. 


\noindent\textbf{Impact of Losses: }
To understand which technique contributes most to the incremental learning performance, we created different variants of the approach by selecting different losses. Table~\ref{tab:loss} presents the results at different incremental steps and the average accuracy for incremental learning. We can observe that without exemplars (first two rows), the accuracy drops substantially (by 40\%) compared to the optimal performance. Knowledge distillation merely improves the accuracy by 3\% (row 2-row 1, or row 4-row 3). Overall, the performance gain is dominated by the use of exemplars and knowledge distillation provides limited accuracy enhancement.  

\begin{table}[]
\caption{Effect of different losses on the incremental learning performance. CE refers to Cross Entropy, KD refers to Knowledge Distillation, N refers to New samples, and O refers to Old samples (exemplars). }
\vspace{0.05in}
\resizebox{0.49\textwidth}{!}{
\begin{tabular}{|l|l|l|l|l|l|l|l|}
\hline
\# of class                                                        & 5     & 8     & 11    & 14    & 17    & 20    & Avg            \\ \hline
CE\_N                                                              & 96.97 & 60.23 & 43.79 & 35.73 & 29.46 & 26.22 & \textbf{39.09} \\ \hline
CE\_N+KD\_N                                                        & 97.08 & 60.75 & 43.34 & 37.43 & 36.20 & 34.85 & \textbf{42.52} \\ \hline
CE\_N+CE\_O                                                        & 97.25 & 88.77 & 79.07 & 72.88 & 72.82 & 57.27 & \textbf{74.16} \\ \hline
\begin{tabular}[c]{@{}l@{}}CE\_N+CE\_O\\ +KD\_N\end{tabular}       & 96.78 & 84.65 & 78.27 & 77.91 & 73.60 & 72.55 & \textbf{77.39} \\ \hline
\begin{tabular}[c]{@{}l@{}}CE\_N+CE\_O\\ +KD\_O\end{tabular}       & 97.12 & 85.72 & 80.64 & 78.99 & 74.30 & 71.93 & \textbf{78.32} \\ \hline
\begin{tabular}[c]{@{}l@{}}CE\_N+CE\_O+\\ KD\_N+KD\_O\end{tabular} & 97.35 & 87.92 & 81.47 & 77.66 & 73.80 & 73.27 & \textbf{78.82} \\ \hline
\end{tabular}
}

\vspace{-0.2in}
\label{tab:loss}
\end{table}

\section{Final Remarks}
\label{sec:majhead}

In this work, we first identified that a more transferable feature representation of the base model might be beneficial for incremental learning. Then, we introduced multitask learning to the base model training stage to improve the generalizability of representations. With two audio datasets, we explored the impact of multitask creation, exemplar quantity, and different losses. The results show that our approach improves the average incremental accuracy by up to 5.5\%. A thorough comparison with more baselines is planed as a future work. Our work opens the door to improving the quality of base model in incremental learning, which motivates the exploration of various generalization techniques in the future.


\section{Acknowledgement}
This work is partially supported by Nokia Bell Labs through their donation for the Centre of Mobile, Wearable Systems and Augmented Intelligence to the University of Cambridge. This work is also supported by ERC through Project 833296 (EAR). 


\bibliographystyle{IEEEbib}
\bibliography{refs}

\begin{thebibliography}{10}

\bibitem{chen2014small}
Guoguo Chen, Carolina Parada, and Georg Heigold,
\newblock ``Small-footprint keyword spotting using deep neural networks,''
\newblock in {\em 2014 IEEE International Conference on Acoustics, Speech and
  Signal Processing (ICASSP)}. IEEE, 2014, pp. 4087--4091.

\bibitem{szoke2005comparison}
Igor Sz{\"o}ke, Petr Schwarz, Pavel Matejka, Luk{\'a}s Burget, Martin
  Karafi{\'a}t, Michal Fapso, and Jan Cernock{\`y},
\newblock ``Comparison of keyword spotting approaches for informal continuous
  speech.,''
\newblock in {\em Interspeech}, 2005, pp. 633--636.

\bibitem{wang2014face}
Weihong Wang, Jie Yang, Jianwei Xiao, Sheng Li, and Dixin Zhou,
\newblock ``Face recognition based on deep learning,''
\newblock in {\em International Conference on Human Centered Computing}.
  Springer, 2014, pp. 812--820.

\bibitem{diethe2019continual}
Tom Diethe, Tom Borchert, Eno Thereska, Borja Balle, and Neil Lawrence,
\newblock ``Continual learning in practice,''
\newblock {\em arXiv preprint arXiv:1903.05202}, 2019.

\bibitem{van2019three}
Gido~M Van~de Ven and Andreas~S Tolias,
\newblock ``Three scenarios for continual learning,''
\newblock {\em arXiv preprint arXiv:1904.07734}, 2019.

\bibitem{aljundi2018memory}
Rahaf Aljundi, Francesca Babiloni, Mohamed Elhoseiny, Marcus Rohrbach, and
  Tinne Tuytelaars,
\newblock ``Memory aware synapses: Learning what (not) to forget,''
\newblock in {\em Proceedings of the European Conference on Computer Vision
  (ECCV)}, 2018, pp. 139--154.

\bibitem{kirkpatrick2017overcoming}
James Kirkpatrick, Razvan Pascanu, Neil Rabinowitz, Joel Veness, Guillaume
  Desjardins, Andrei~A Rusu, Kieran Milan, John Quan, Tiago Ramalho, Agnieszka
  Grabska-Barwinska, et~al.,
\newblock ``Overcoming catastrophic forgetting in neural networks,''
\newblock {\em Proceedings of the national academy of sciences}, vol. 114, no.
  13, pp. 3521--3526, 2017.

\bibitem{zenke2017continual}
Friedemann Zenke, Ben Poole, and Surya Ganguli,
\newblock ``Continual learning through synaptic intelligence,''
\newblock in {\em International Conference on Machine Learning}. PMLR, 2017,
  pp. 3987--3995.

\bibitem{li2017learning}
Zhizhong Li and Derek Hoiem,
\newblock ``Learning without forgetting,''
\newblock {\em IEEE transactions on pattern analysis and machine intelligence},
  vol. 40, no. 12, pp. 2935--2947, 2017.

\bibitem{wu2019large}
Yue Wu, Yinpeng Chen, Lijuan Wang, Yuancheng Ye, Zicheng Liu, Yandong Guo, and
  Yun Fu,
\newblock ``Large scale incremental learning,''
\newblock in {\em Proceedings of the IEEE/CVF Conference on Computer Vision and
  Pattern Recognition}, 2019, pp. 374--382.

\bibitem{hou2019learning}
Saihui Hou, Xinyu Pan, Chen~Change Loy, Zilei Wang, and Dahua Lin,
\newblock ``Learning a unified classifier incrementally via rebalancing,''
\newblock in {\em Proceedings of the IEEE/CVF Conference on Computer Vision and
  Pattern Recognition}, 2019, pp. 831--839.

\bibitem{zhao2020maintaining}
Bowen Zhao, Xi~Xiao, Guojun Gan, Bin Zhang, and Shu-Tao Xia,
\newblock ``Maintaining discrimination and fairness in class incremental
  learning,''
\newblock in {\em Proceedings of the IEEE/CVF Conference on Computer Vision and
  Pattern Recognition}, 2020, pp. 13208--13217.

\bibitem{hinton2015distilling}
Geoffrey Hinton, Oriol Vinyals, and Jeff Dean,
\newblock ``Distilling the knowledge in a neural network,''
\newblock {\em arXiv preprint arXiv:1503.02531}, 2015.

\bibitem{zhou2019m2kd}
Peng Zhou, Long Mai, Jianming Zhang, Ning Xu, Zuxuan Wu, and Larry~S Davis,
\newblock ``M2kd: Multi-model and multi-level knowledge distillation for
  incremental learning,''
\newblock {\em arXiv preprint arXiv:1904.01769}, 2019.

\bibitem{jha2021continual}
Saurav Jha, Martin Schiemer, Franco Zambonelli, and Juan Ye,
\newblock ``Continual learning in sensor-based human activity recognition: an
  empirical benchmark analysis,''
\newblock {\em arXiv preprint arXiv:2104.09396}, 2021.

\bibitem{lopez2017gradient}
David Lopez-Paz and Marc'Aurelio Ranzato,
\newblock ``Gradient episodic memory for continual learning,''
\newblock {\em Advances in neural information processing systems}, vol. 30, pp.
  6467--6476, 2017.

\bibitem{hayes2020remind}
Tyler~L Hayes, Kushal Kafle, Robik Shrestha, Manoj Acharya, and Christopher
  Kanan,
\newblock ``Remind your neural network to prevent catastrophic forgetting,''
\newblock in {\em European Conference on Computer Vision}. Springer, 2020, pp.
  466--483.

\bibitem{iscen2020memory}
Ahmet Iscen, Jeffrey Zhang, Svetlana Lazebnik, and Cordelia Schmid,
\newblock ``Memory-efficient incremental learning through feature adaptation,''
\newblock in {\em European Conference on Computer Vision}. Springer, 2020, pp.
  699--715.

\bibitem{mittal2021essentials}
Sudhanshu Mittal, Silvio Galesso, and Thomas Brox,
\newblock ``Essentials for class incremental learning,''
\newblock in {\em Proceedings of the IEEE/CVF Conference on Computer Vision and
  Pattern Recognition}, 2021, pp. 3513--3522.

\bibitem{caruana1997multitask}
Rich Caruana,
\newblock ``Multitask learning,''
\newblock {\em Machine learning}, vol. 28, no. 1, pp. 41--75, 1997.

\bibitem{liu2020mnemonics}
Yaoyao Liu, Yuting Su, An-An Liu, Bernt Schiele, and Qianru Sun,
\newblock ``Mnemonics training: Multi-class incremental learning without
  forgetting,''
\newblock in {\em Proceedings of the IEEE/CVF conference on Computer Vision and
  Pattern Recognition}, 2020, pp. 12245--12254.

\bibitem{castro2018end}
Francisco~M Castro, Manuel~J Mar{\'\i}n-Jim{\'e}nez, Nicol{\'a}s Guil, Cordelia
  Schmid, and Karteek Alahari,
\newblock ``End-to-end incremental learning,''
\newblock in {\em Proceedings of the European conference on computer vision
  (ECCV)}, 2018, pp. 233--248.

\bibitem{rebuffi2017icarl}
Sylvestre-Alvise Rebuffi, Alexander Kolesnikov, Georg Sperl, and Christoph~H
  Lampert,
\newblock ``icarl: Incremental classifier and representation learning,''
\newblock in {\em Proceedings of the IEEE conference on Computer Vision and
  Pattern Recognition}, 2017, pp. 2001--2010.

\bibitem{goodfellow2014generative}
Ian Goodfellow, Jean Pouget-Abadie, Mehdi Mirza, Bing Xu, David Warde-Farley,
  Sherjil Ozair, Aaron Courville, and Yoshua Bengio,
\newblock ``Generative adversarial nets,''
\newblock {\em Advances in neural information processing systems}, vol. 27,
  2014.

\bibitem{multi_task_survey_9392366}
Yu~Zhang and Qiang Yang,
\newblock ``A survey on multi-task learning,''
\newblock {\em IEEE Transactions on Knowledge and Data Engineering}, pp. 1--1,
  2021.

\bibitem{salamon2014dataset}
Justin Salamon, Christopher Jacoby, and Juan~Pablo Bello,
\newblock ``A dataset and taxonomy for urban sound research,''
\newblock in {\em Proceedings of the 22nd ACM international conference on
  Multimedia}, 2014, pp. 1041--1044.

\bibitem{kwon2021fasticarl}
Young~D Kwon, Jagmohan Chauhan, and Cecilia Mascolo,
\newblock ``Fasticarl: Fast incremental classifier and representation learning
  with efficient budget allocation in audio sensing applications,''
\newblock {\em arXiv preprint arXiv:2106.07268}, 2021.

\end{thebibliography}

\end{document}